\documentclass[runningheads]{llncs}

\usepackage[mobile]{eccv}

\usepackage{eccvabbrv}

\usepackage{graphicx}
\usepackage{booktabs}

\usepackage[accsupp]{axessibility}  %

\usepackage[pagebackref,breaklinks,colorlinks,linkcolor=blue,citecolor=eccvblue,filecolor=magenta,urlcolor=blue]{hyperref}

\usepackage{orcidlink}
\usepackage{graphicx} %
\usepackage{arydshln} %
\usepackage{amssymb} 
\usepackage{booktabs}
\usepackage{amsmath}
\usepackage[numbers]{natbib}
\usepackage[table]{xcolor}
\usepackage{subcaption}
\usepackage{hyperref}
\usepackage{multirow} 
\usepackage{mdframed}
\usepackage[percent]{overpic}
\usepackage{tikz}
\usepackage{enumitem}
\usepackage{wrapfig}
\usetikzlibrary{calc}

\definecolor{eloiblue}{RGB}{0,0,190}
\definecolor{pltblue}{RGB}{31,119,180}
\definecolor{pltred}{RGB}{214,39,40}

\usepackage{pgfplots}
\pgfplotsset{compat=1.18}

\begin{document}

\title{How Far Can 5,500 Hours of Driving Take You? A Scaling Law Analysis of Video Diffusion Models} 

\titlerunning{A Scaling Law Analysis of Video Diffusion Models}

\author{Victor Besnier \and
Anh-Quan Cao \and
Elias Ramzi \and
Spyros Gidaris \and
Tuan-Hung Vu \and
Andrei Bursuc \and
Eloi Zablocki \and
Matthieu Cord}

\authorrunning{V.~Besnier et al.}

\institute{valeo.ai, Paris, France \\[1em]
Project page: \url{https://valeoai.github.io/VATIX/}
}

\maketitle

\vspace{-0.5cm}
\begin{abstract}
  Video generation for autonomous driving cannot follow the web-scale route: driving data is expensive to collect, bound by privacy requirements, and cannot be scraped at will, so models must make the most of a fixed corpus.
  We present a systematic scaling-law study of video diffusion models trained from scratch on driving data: a family of models from 1M to 9B parameters, trained at different exposures on up to 5,500 hours of driving.
  Validation loss follows consistent power laws in both model size and training exposure, answering the questions that shape a training budget: whether compute is better spent on longer training or on a larger model, and whether more data is needed.
  Loss improves much faster with training exposure than with model size, making longer training the most effective way to improve a fixed model under limited compute. However, larger models continue to achieve lower asymptotic loss, so compute-optimal scaling still favors increasing model size when sufficient compute and data are available.
  Guided by these laws, we train a 9B-parameter model, to our knowledge the largest video diffusion model trained from scratch on driving data: it sets a new open-source state of the art for driving video generation, as measured on nuScenes.
  Our code and pretrained models are available at \url{https://github.com/valeoai/VATIX}. 
  NATIX is separately releasing the underlying driving data in stages.

  \keywords{Video Diffusion Models \and Scaling Laws \and Driving Dataset}
\end{abstract}

\section{Introduction}\label{sec:intro}

Video generation models can now create footage that looks completely real, a breakthrough driven almost entirely by scale~\citep{ho2020denoising,lipman2023flow,esser2024scaling}.
For autonomous driving, this capability is invaluable.
It can generate rare, dangerous corner cases that a real test fleet could never safely capture.
When these models are trained to predict how an environment reacts to the actions of the ego-car, they become \emph{world models}, providing a rich simulation space for downstream motion prediction and planning~\citep{bartoccioni2025vavim,xu2026eponav2,tu2025wm_survey}.
However, driving data cannot be scraped freely from the web: high collection costs and strict privacy laws leave most labs with a fixed dataset and a limited number of GPUs.
This brings us to a concrete question: how far can a fixed dataset, 5\,500 hours of driving video, actually take a video diffusion model?

To answer this, we need to know exactly where each unit of compute is best spent: is it better to build a larger model, or to train for more steps? In large language models (LLMs), \emph{scaling laws} provide empirical formulae to answer this question~\citep{kaplan2020scaling,hoffmann2022training}.
However, it is entirely unclear if these principles transfer to video diffusion models.
Two differences separate our setting from these studies.
First, the optimization objective relies on denoising or flow matching rather than next-token prediction.
Second, we operate in the data-constrained regime, unlike LLMs whose text corpora are practically unbounded. In Chinchilla-style scaling~\citep{hoffmann2022training}, growing the data axis means sourcing more \emph{unique} tokens, since web-scale text rarely needs repeating. Here, the driving corpus is fixed at 5,500 hours: far larger than any public driving dataset, yet still small enough that repeated \emph{epochs} over the same clips are unavoidable. Training exposure, the total number of samples seen, unique or repeated alike, thus becomes the primary scalable resource, distinct from dataset size itself.
This motivates an empirical study of scaling laws for video diffusion models under these constraints.
Following prior work on LLMs, we fit a simple asymptotic power law, $\mathcal{L}(x) = L_0 + A\, x^{-\alpha}$, to the validation loss along three complementary axes: model scaling ($x := N$, the number of parameters), training scaling ($x := D$, the training exposure), and compute scaling ($x := C$, the compute budget).

We proceed in two steps: first we fit the scaling laws, then we spend the budget where they point.
The fits are accurate across more than 200 runs spanning 1.6M to 1.1B parameters, and three findings follow.
\textbf{(1)} The training-exposure exponent is far steeper than the model-size exponent, so longer training is the fastest way to improve a fixed model, while the asymptotic loss keeps falling with capacity, so compute-optimal budgets still favor larger models.
\textbf{(2)} The fixed corpus does not break the laws: within the fewer than five epochs we run, repeating the data behaves like fresh samples~\citep{prabhudesai2026diffusion,muennighoff2023scaling}, so the 5,500 hours are not yet the bottleneck. We hypothesize this is because flow matching resamples noise and timestep at every pass, unlike the fixed targets of autoregressive training.
And \textbf{(3)} the laws extrapolate: a single 9B-parameter model, to our knowledge the largest open-source video diffusion model trained from scratch on driving data, reaches a validation loss predicted before training to within 3.6\% of a law extrapolated 8$\times$ beyond the largest fitted model, the lowest loss among all our models.
To enable action control and fair comparison with baselines, which all rely on trajectory conditioning, we further post-finetune it with pseudo-trajectory conditioning. The resulting model sets a new open-source state of the art for driving video generation on nuScenes.

\section{Related Work}\label{sec:related_works}

\paragraph{Driving Diffusion Models.}

Recent driving world models generate future driving scenes using autoregressive or trajectory-conditioned video generation, demonstrating that large-scale video pretraining learns transferable representations for motion prediction and planning \citep{zhang2025epona, bartoccioni2025vavim}. More recent methods improve controllability and physical consistency through multimodal conditioning, such as depth, ego-motion, and object dynamics, while larger foundation models further benefit from scaling data and model capacity \citep{hassan2025gem, rahimi2026mad, russell2025gaia, ali2025world}.

Most of these approaches adapt general-purpose video diffusion backbones pretrained on web-scale data using parameter-efficient tuning methods such as LoRA~\citep{hu2022lora}. Models such as Sora~\citep{openai2024sora}, Wan~\citep{wan2025wan}, and LTX~\citep{hacohen2024ltx} achieve high visual quality, but their proprietary pretraining prevents controlled analysis of how performance scales with driving-specific data and compute.

We instead train diffusion models from scratch on 5,500 hours of driving video, roughly three times as much public data as used by VISTA~\citep{gao2024vista} and GEM~\citep{hassan2025gem}, enabling a controlled study of scaling with model size, data exposure, and compute. Unlike adaptation-based works that optimize transfer from web-scale priors targeting video quality, our goal is to characterize these scaling laws for driving video generation. The resulting laws are useful to both lines of work when allocating a driving-video budget.

\paragraph{Scaling Laws.}

Scaling laws were first established in the context of large language models, showing that model performance follows predictable power-law relationships with respect to model size, dataset size, and compute \citep{kaplan2020scaling}. Subsequent work refined these findings by demonstrating that optimal performance is achieved when model capacity and data are scaled jointly under a fixed compute budget, leading to the compute-optimal training paradigm known as Chinchilla \citep{hoffmann2022training}. Extensions to data-constrained regimes further highlighted that these optimal trade-offs change when training is limited by dataset availability rather than compute, emphasizing the importance of distinguishing between compute- and data-limited settings \citep{muennighoff2023scaling}.

Beyond language modeling, recent work has begun exploring scaling laws in other modalities. In particular, diffusion and rectified flow models exhibit stable, predictable power-law scaling in image synthesis \citep{liang2026scaling, esser2024scaling}. \cite{yin2025scaling} studies scaling laws for video diffusion models, but relies on pretrained models, which may bias the measured scaling behavior. This trends that also hold across multimodal text-image architectures \citep{shukor2025scaling}. Concurrently, diffusion models have proven highly effective in data-constrained regimes, frequently outperforming autoregressive alternatives when training data is limited \citep{prabhudesai2026diffusion}.

Closest in spirit, scaling laws for pretraining agents and world models~\citep{pearce2025scaling} target only game environments and autoregressive architectures, while~\citep{Nauman2025data} target end-to-end driving. Finally, VaViM \citep{bartoccioni2025vavim} derives scaling laws for autoregressive video models trained on driving data, but these do not directly transfer to diffusion and flow matching, whose training objective and data regime differ: scaling laws for diffusion-based video generation remain missing.

\section{Method and Experimental Setup}\label{sec:methods}

\subsection{Conditional Flow Matching (CFM)}
We formulate video generation as learning a transport map from a simple prior distribution $x_0 \sim \mathcal{N}(0, I)$ to the data distribution $x_1 \sim p_{\text{data}}$, where each sample $x \in \mathbb{R}^{F \times C \times H \times W}$ represents a sequence of frames. Conditional Flow Matching learns a time-dependent velocity field that defines a continuous transport between these distributions \citep{lipman2023flow}. Given paired samples $(x_0, x_1)$ and a time $t \in [0,1]$, we define a probability path which linearly interpolates between noise and data:
\begin{equation}
x_t = (1 - t)x_0 + t x_1.
\end{equation}

The corresponding target velocity field defines the direction of transport along the interpolation path. It is given by
\begin{equation}
u(x_t, t) = \frac{d x_t}{dt} = x_1 - x_0.
\end{equation}

A neural network $u_\theta(x_t, t, c)$ is trained to regress this velocity using:
\begin{equation}
\mathcal{L} = \mathbb{E}_{x_0, x_1, t} \left[ \| u_\theta(x_t, t, c) - (x_1 - x_0) \|^2 \right].
\label{eq:cfm_loss}
\end{equation}
Here $c$ is the conditioning signal of the model: the first video frame for image-to-video generation, later extended with the ego-trajectory (\autoref{subsec:visual_quality}).

At inference time, samples are generated by solving the ordinary differential equation \( \frac{dx}{dt} = u_\theta(x, t, c) \), which is numerically integrated using standard solvers such as Euler or higher-order methods to transform noise into data.

\subsection{Model architecture}\label{subsec:impl}

\begin{figure}[t]
    \centering
    \includegraphics[width=0.8\linewidth]{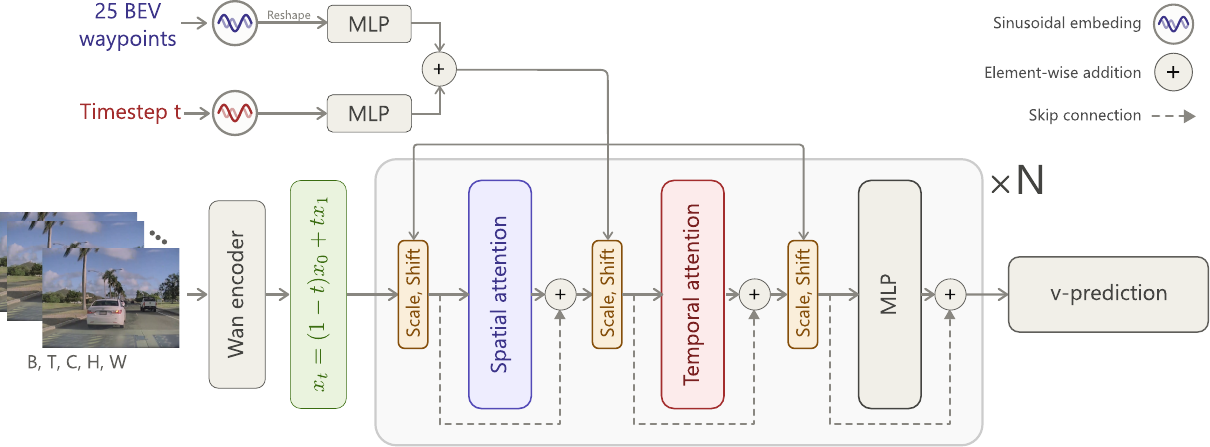}
    \vspace{-1em}
   \caption{\textbf{Model architecture.} Video latents from the Wan encoder are noised and processed by $N$ transformer blocks, each containing spatial attention, temporal attention, and an MLP, modulated by AdaLN. The timestep $t$ and ego-trajectory waypoints are embedded with sinusoidal features, and each is passed through an MLP. They are then summed and chunked into scale and shift. The output is the velocity (v-prediction).}
    \label{fig:arch}
    \vspace{-0.5cm}
\end{figure}

Based on the Diffusion Transformer (DiT)~\cite{peebles2023dit} (\autoref{fig:arch}), our architecture is a spatio-temporal video transformer operating on latents from an off-the-shelf Wan~2.1 VAE \citep{wan2025wan}. These input latents are augmented with spatial and temporal positional embeddings, then processed through a stack of transformer blocks. Each block applies spatial and temporal self-attention followed by a feed-forward layer, all modulated by adaptive LayerNorm (AdaLN)~\cite{peebles2023dit} conditioned on a timestep sinusoidal embedding. Scaling this architecture from 1.6M to 9B parameters yields the 12 model sizes listed in \autoref{tab:model-size}.

\begin{table}[t]
\centering
\caption{\textbf{Model size and training compute.} * denotes the target model size. Training TFLOPs are estimated with \texttt{thop} as three times the FLOPs of a forward pass, approximating one training step (one forward and two backward-pass equivalents).}
\vspace{-10pt}
\resizebox{\linewidth}{!}{%
\begin{tabular}{lcccccccccccc}
\toprule
Model &
Pico & Nano & Micro & Tiny & Small & Medium & Base & Big &
Large & XLarge & 1B & 9B* \\
\midrule
Params (M) &
1.6 & 3.8 & 9.2 & 19.7 & 33.8 & 63.3 & 134.9 & 295.5 &
447.3 & 643.1 & 1142.9 & 9096.4 \\
TFLOPs &
0.02 & 0.05 & 0.14 & 0.30 & 0.61 & 1.10 & 2.47 & 5.61 &
8.79 & 12.99 & 23.09 & 143.84 \\
\bottomrule
\end{tabular}%
}

\label{tab:model-size}
\end{table}

We train our image-to-video flow-matching model with DDP for models with $\leq$1B parameters and with FSDP2 for the 9B model. The model is trained in bfloat16 precision with the AdamW optimizer, sweeping the global batch size over $\{16, 32, 64\}$. The learning rate is swept over $\{10^{-4}, 5 \times 10^{-5}, 10^{-5}\}$ for models below 1B parameters and set to $10^{-5}$ above, and remains constant after warmup. Weight decay is 0.01 and gradients are clipped at 1.0.

Training uses a v-prediction formulation, whose target coincides with the velocity-regression objective of \autoref{eq:cfm_loss} under the linear interpolation path, with a shifted log-normal noise schedule ($\mu = -0.7$, $\sigma = 1.4$), which biases training toward higher-noise regimes. Each run includes a 25k-step learning rate warmup. Exponential moving average (EMA) is applied with a decay of 0.9999.

The model is trained on 25-frame front-view windows at a resolution of $320 \times 416$, derived from NATIX footage (\autoref{subsec:dataset}). The input to the Wan~2.1 encoder has shape $25 \times 3 \times 320 \times 416$, where 3 denotes the RGB channels. The encoder maps each clip to a latent representation of shape $7 \times 16 \times 40 \times 52$, comprising 7 latent frames obtained via temporal downsampling (factor 4 while preserving the first frame), 16 channels, and a spatial resolution of $40 \times 52$.

\subsection{Dataset}\label{subsec:dataset}
We train our models on a large-scale, multi-camera driving dataset comprising approximately 5,500 hours of raw footage, collected between June and December 2025, segmented into 307K videos of approximately 1 min each and spanning 28 countries across Europe, North America, and Japan. The data is provided by NATIX.
Independently of this work, NATIX is releasing this dataset in stages: a first 100-hour subset is already public, with an initial target of 2,000+ hours and additional releases expected over time.
We split each video into 2.5-second clips at 9\,Hz. We retain only the front-facing camera stream. The footage is anonymized: license plates and faces are blurred. Train/val/test splits are performed at the trip level. We report results using an IID split (85/5/10), the split is constructed so that the distribution of countries is preserved. This yielding 260,853 training videos, 15,404 validation videos, and 30,822 test videos. The training set thus contains around 6.3M distinct 2.5-second clips, so the exposures used in \autoref{sec:scaling} (10M--28M samples) correspond to roughly 1.6--4.5 epochs.

\section{Deriving Scaling Laws}\label{sec:scaling}

Our objective is to derive scaling laws for flow-matching video generation models in the context of driving scenes. These laws tell us where the next unit of compute is best spent: on more training steps, on a larger model, or on more data. They also let us plan the training of a flagship model: they estimate the validation loss that a 9-billion-parameter model (the largest trainable on a node of 4$\times$H100 80GB with FSDP2) can reach, along with the number of training epochs required to reach it. This goal structures our study along three axes: model scaling~(\autoref{subsec:N}), training scaling~(\autoref{subsec:D}), and compute scaling~(\autoref{subsec:C}). Each subsection follows the same structure: the goal, the experimental setup, the results, and a boxed prediction for the 9B run.

All three axes are fit with the same asymptotic power law:
\begin{equation}
\mathcal{L}(x) = L_0 + A\, x^{-\alpha},
\label{eq:scaling_law}
\end{equation}
where $x$ denotes the scaling variable ($N$, $D$, or $C$) and $\mathcal{L}(x)$ the corresponding validation loss; $L_0$ is the fitted asymptotic loss, $A$ sets the magnitude of the decay, i.e., the size of the initial gap, and $\alpha$ sets the rate at which the loss decreases with $x$. We subscript the exponent by its axis, $\alpha_N$, $\alpha_D$, or $\alpha_C$, whenever the axis is not clear from context.

\subsection{\texorpdfstring{\textbf{Model scaling} ($x{=}N$)}{Model scaling (x = N)}}\label{subsec:N}
\paragraph{\textbf{Goal.}} 
We investigate how the validation loss scales with model size \(N\) under fixed training exposure, assessing whether this empirical behavior follows the power law of \autoref{eq:scaling_law} and whether it can be reliably extrapolated to larger models.

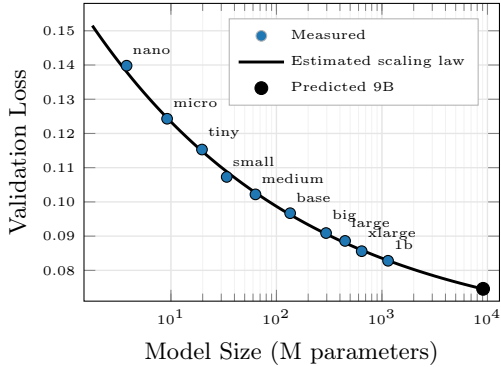
\begin{figure}[t]
\centering

\begin{subfigure}[b]{0.38\textwidth}
\centering
\begin{tabular}{lrc}
\hline
Model & Params (M) & Val Loss \\
\hline
Pico   & 1.6    & 0.1528 \\
Nano   & 3.8    & 0.1398 \\
Micro  & 9.2    & 0.1243 \\
Tiny   & 19.7   & 0.1153 \\
Small  & 33.8   & 0.1073 \\
Medium & 63.3   & 0.1022 \\
Base   & 134.9  & 0.0967 \\
Big    & 295.5  & 0.0909 \\
Large  & 447.3  & 0.0886 \\
XLarge & 643.1  & 0.0856 \\
1B     & 1142.9 & 0.0828 \\
\hline
\end{tabular}
\caption{Observed validation loss at 10M samples seen across model sizes.}
\label{tab:scaling_model_size}
\end{subfigure}
\hfill
\begin{subfigure}[b]{0.58\textwidth}
\centering
\begin{tikzpicture}
\begin{axis}[
    width=\linewidth,
    height=0.78\linewidth,
    xmode=log,
    xmin=1.5, xmax=13000,
    ymin=0.071, ymax=0.158,
    xlabel={Model Size (M parameters)},
    ylabel={Validation Loss},
    xtick={10, 100, 1000, 10000},
    ytick={0.08, 0.09, 0.10, 0.11, 0.12, 0.13, 0.14, 0.15},
    yticklabels={0.08, 0.09, 0.10, 0.11, 0.12, 0.13, 0.14, 0.15},
    grid=both,
    major grid style={line width=0.5pt, draw=gray!20},
    minor grid style={line width=0.3pt, draw=gray!10},
    tick label style={font=\tiny},
    label style={font=\scriptsize},
    xlabel near ticks,
    ylabel near ticks,
    legend cell align={left},
    legend style={at={(0.95,0.95)}, anchor=north east, font=\tiny, row sep=1pt, draw=gray!60}
]

\addplot[
    only marks,
    mark=*,
    mark options={fill=pltblue, draw=none},
    mark size=2pt
] coordinates {
    (3.8, 0.1398)
    (9.2, 0.1243)
    (19.7, 0.1153)
    (33.8, 0.1073)
    (63.3, 0.1022)
    (134.9, 0.0967)
    (295.5, 0.0909)
    (447.3, 0.0886)
    (643.1, 0.0856)
    (1142.9, 0.0828)
};
\addlegendentry{Measured}

\addplot[
    black,
    line width=1.1pt,
    domain=1.8:10000,
    samples=300
] {0.059567 + 0.104156 * x^(-0.212599)};
\addlegendentry{Estimated scaling law}

\addplot[
    only marks,
    mark=*,
    mark options={fill=black, draw=black},
    mark size=2.4pt
] coordinates {
    (9096.4, 0.0746)
};
\addlegendentry{Predicted 9B}

\node[above right, font=\tiny, xshift=-1pt, yshift=1pt] at (axis cs:3.8, 0.1398) {nano};
\node[above right, font=\tiny, xshift=-1pt, yshift=1pt] at (axis cs:9.2, 0.1243) {micro};
\node[above right, font=\tiny, xshift=-1pt, yshift=1pt] at (axis cs:19.7, 0.1153) {tiny};
\node[above right, font=\tiny, xshift=-1pt, yshift=1pt] at (axis cs:33.8, 0.1073) {small};
\node[above right, font=\tiny, xshift=-1pt, yshift=1pt] at (axis cs:63.3, 0.1022) {medium};
\node[above right, font=\tiny, xshift=-1pt, yshift=1pt] at (axis cs:134.9, 0.0967) {base};
\node[above right, font=\tiny, xshift=-1pt, yshift=1pt] at (axis cs:295.5, 0.0909) {big};
\node[above right, font=\tiny, xshift=-1pt, yshift=1pt] at (axis cs:447.3, 0.0886) {large};
\node[above right, font=\tiny, xshift=-1pt, yshift=1pt] at (axis cs:643.1, 0.0856) {xlarge};
\node[above right, font=\tiny, xshift=-1pt, yshift=1pt] at (axis cs:1142.9, 0.0828) {1b};

\end{axis}
\end{tikzpicture}
\caption{Scaling-law extrapolation to 9B at 10M samples seen ($\mathcal{L}(N) = 0.0596 + 0.1042 \cdot N^{-0.2126}$).}
\label{fig:scaling_model_size_b}
\end{subfigure}
\vspace{-1em}
\caption{\textbf{Model scaling analysis.} Validation loss consistently decreases with increasing model size, and the fitted scaling law extrapolates this trend to larger models.}
\label{fig:scaling_model_size}
\vspace{-15pt}
\end{figure}

\paragraph{\textbf{Setup.}} 
We train 72 models spanning 11 model sizes (\autoref{tab:model-size}). For each size, we evaluate multiple learning rates, batch sizes, and random seeds while keeping the training exposure fixed at 10M samples. We report the best validation loss for each model size \autoref{tab:scaling_model_size} and fit \autoref{eq:scaling_law} to the resulting measurements:
\begin{equation}
    \mathcal{L}(N) = 0.0595 + 0.1041 \cdot N^{-0.2125},
    \label{eq:model_scaling_fit_N}
\end{equation}
where \(N\) is the number of parameters and the fitted exponent is \(\alpha_N = 0.2125\).

\paragraph{\textbf{Results.}} 
The fitted law closely matches the measured validation losses, achieving an RMSE of \(8.42 \times 10^{-4}\) and a MAPE (Mean Absolute Percentage Error) of \(0.56\%\), accurately capturing the dependence on model size. The extrapolated scaling curve is shown in \autoref{fig:scaling_model_size_b}, revealing a clear power law regime with diminishing returns as model size increases. For example, \autoref{eq:model_scaling_fit_N} predicts that scaling from \(134.9\)M to \(447.3\)M parameters reduces validation loss from \(0.0963\) to \(0.0880\), while further scaling to \(1.1\)B parameters reaches \(0.0829\). Despite the diminishing returns, no saturation is observed up to the largest trained model (\(1.1\)B parameters), suggesting that further scaling remains beneficial.

\vspace{35pt}

\begin{mdframed}[linewidth=0.8pt]
\textbf{Prediction for the 9B run.}
The fitted scaling law remains accurate over more than three orders of magnitude in model size (\(1.6\)M–\(1.1\)B parameters). Extrapolating to a \(9\)B-parameter model trained with the same exposure (\(D \approx 10^7\) samples) predicts a validation loss of
\[
\mathcal{L}(9\mathrm{B}) \approx 0.0746.
\]
This estimate suggests that substantial gains remain achievable through further model scaling beyond the largest model considered in this study.
\end{mdframed}

\subsection{\texorpdfstring{\textbf{Training scaling} ($x{=}D$)}{Training scaling (x = D)}}\label{subsec:D}

\paragraph{\textbf{Goal.}} 
We investigate how validation loss scales with training exposure \(D\), i.e., the number of samples seen during training, while keeping model size fixed. We assess whether the loss evolution follows the power law of \autoref{eq:scaling_law} and whether this learned scaling behavior remains consistent across model sizes.

\begin{figure}[t]
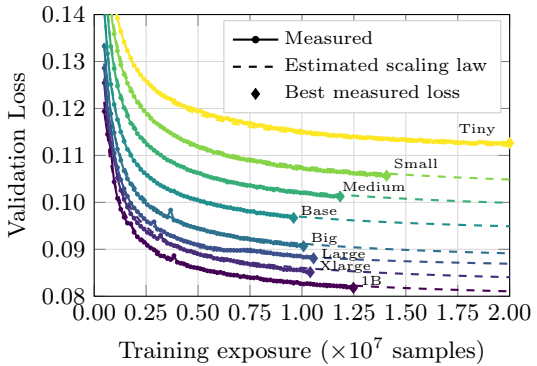

\centering

\begin{subfigure}[b]{0.38\textwidth}
\centering
\begin{tabular}{lccc}
\toprule
Model & $L_0$ & $A$ & $\alpha_D$ \\
\midrule
Pico & 0.147 & 0.006056 & 0.740 \\
Nano & 0.131 & 0.006338 & 0.729 \\
Micro & 0.120 & 0.004406 & 0.830 \\
Tiny & 0.108 & 0.006601 & 0.707 \\
Small & 0.100 & 0.007710 & 0.665 \\
Medium & 0.096 & 0.006349 & 0.705 \\
Base & 0.092 & 0.004922 & 0.758 \\
Big & 0.086 & 0.005212 & 0.731 \\
Large & 0.085 & 0.003458 & 0.842 \\
XLarge & 0.081 & 0.005055 & 0.723 \\
1B & 0.078 & 0.004997 & 0.712 \\

\bottomrule
\end{tabular}
\caption{Parameters of the fitted training scaling law
$\mathcal{L}(D)=L_0+A D^{-\alpha_D}$ across model sizes, where \(D\) denotes the training exposure in units of \(10^7\).}
\label{tab:scaling_laws_training_exp}
\end{subfigure}
\hfill
\begin{subfigure}[b]{0.58\textwidth}
\centering
%
\caption{Per-model training curves and fitted scaling laws.}
\label{fig:epoch_fit_convergence_multiscale_b}
\end{subfigure}
\caption{\textbf{Training dynamics per model size.} Solid curves show measured validation loss during training; dashed curves show the corresponding per-model power law fit, extrapolated across the full exposure range shown; diamonds mark the best measured loss for each model. The dynamics are consistent across scales and well captured by per-model fits, while larger models converge to lower loss.}
\vspace{-0.5cm}
\label{fig:epoch_fit_convergence_multiscale}
\end{figure}

\paragraph{\textbf{Setup.}} Our grid search comprises more than 200 runs, spanning learning rates $\{10^{-4}, 5\times10^{-5}, 10^{-5}\}$ and batch sizes $\{16, 32, 64\}$ across all 11 model sizes and training exposures of up to 28M samples.
For each model size and exposure, we retain the best validation loss across hyperparameter configurations and fit the resulting scaling curve $\mathcal{L}(D)$. 
The fitted parameters are reported in \autoref{tab:scaling_laws_training_exp}, and the corresponding training curves are shown in \autoref{fig:epoch_fit_convergence_multiscale_b}.

\paragraph{\textbf{Results.}} 
The fitted scaling laws accurately describe the optimization dynamics, achieving a MAPE below \(1\%\) across model sizes. 
Increasing model size consistently lowers the asymptotic loss \(L_0\), while the convergence exponent \(\alpha_D\) shows no systematic trend across scale (\(0.74\) on average, spread \(0.66\)--\(0.84\)).
As shown by the training curves, models with substantially different capacities follow nearly identical optimization dynamics and converge toward similar diminishing-return regimes. This scale-invariant behavior suggests that long-term training performance can be reliably extrapolated before full convergence, making the fitted scaling law a practical tool for forecasting training outcomes and reducing the cost of large-scale experimentation.

\begin{mdframed}[linewidth=0.8pt]
\textbf{Prediction for the 9B run.}
The training scaling law indicates that optimization exhibits strong diminishing returns, with a stable exponent of $\alpha_D \approx 0.74$ across model sizes. Extrapolating this behavior to the target 9B model suggests that most attainable performance should be reached within a moderate training budget. In practice, we estimate that training up to \(1.5 \times10^7\) samples captures most of the achievable gain: with the mean fitted coefficients of \autoref{tab:scaling_laws_training_exp} ($A \approx 0.005$, $\alpha_D \approx 0.74$), the loss reduction attainable beyond this exposure is below $0.004$.
\end{mdframed}

\paragraph{\textbf{Data restriction ablation.}} To test whether the 5,500-hour corpus is close to a bottleneck, we run a controlled ablation on the Base model: training exposure is held fixed while the pool of unique footage it is drawn from is restricted across four orders of magnitude, from the full 5,500 hours down to 5.5 hours. Restricting the pool forces heavier repetition of the same clips: at a fixed exposure of $1.25\times10^7$ samples, this amounts to close to 200 epochs at the 55-hour threshold and close to 2,000 epochs at the 5.5-hour threshold.

\begin{figure}[t]
\centering
\begin{minipage}[t]{0.62\linewidth}
\centering
\begin{tikzpicture}[baseline=(current bounding box.north)]
\begin{axis}[
    width=\linewidth,
    height=0.7\linewidth,
    xmode=log,
    log basis x=10,
    xmin=3, xmax=10000,
    ymin=0.08, ymax=0.18,
    xtick={5.5, 55, 275, 550, 1375, 5500},
    xticklabels={
        5.5h (0.1\%),
        55h (1\%),
        275h (5\%),
        550h (10\%),
        1375h (25\%),
        5500h (100\%)
    },
    x tick label style={font=\scriptsize, rotate=40, anchor=east},
    ytick={0.08, 0.10, 0.12, 0.14, 0.16, 0.18},
    yticklabels={0.08, 0.10, 0.12, 0.14, 0.16, 0.18},
    y tick label style={font=\scriptsize},
    xlabel={Unique Footage Used (hours)},
    ylabel={Validation Loss},
    label style={font=\scriptsize},
    grid=both,
    grid style={line width=.1pt, draw=gray!20},
    major grid style={line width=.2pt, draw=gray!40},
    legend style={
        at={(0.98,0.95)},
        anchor=north east,
        font=\scriptsize,
        cells={anchor=west},
        row sep=1pt,
        fill=white,
        fill opacity=0.9,
        draw=gray!50
    }
]

\definecolor{color1}{RGB}{253,231,37}
\definecolor{color2}{RGB}{94,201,98}
\definecolor{color3}{RGB}{33,145,140}
\definecolor{color4}{RGB}{59,82,139}
\definecolor{color5}{RGB}{68,1,84}

\addplot[color=color1, mark=*, thick, mark size=1.5pt] coordinates {
    (5.5,0.16911) (55.0,0.09809) (110.0,0.09587) (275.0,0.09347)
    (550.0,0.09654) (1375.0,0.09669) (2750.0,0.09368) (5500.0,0.09392)
};
\addlegendentry{12.5M samples seen}

\addplot[color=color2, mark=square*, thick, mark size=1.5pt] coordinates {
    (5.5,0.16340) (55.0,0.09837) (110.0,0.09679) (275.0,0.09467)
    (550.0,0.09693) (1375.0,0.09704) (2750.0,0.09503) (5500.0,0.09519)
};
\addlegendentry{11.2M samples seen}

\addplot[color=color3, mark=triangle*, thick, mark size=1.8pt] coordinates {
    (5.5,0.14895) (55.0,0.09945) (110.0,0.09906) (275.0,0.09804)
    (550.0,0.09845) (1375.0,0.09863) (2750.0,0.09814) (5500.0,0.09832)
};
\addlegendentry{8.0M samples seen}

\addplot[color=color4, mark=diamond*, thick, mark size=2.0pt] coordinates {
    (5.5,0.13095) (55.0,0.10194) (110.0,0.10183) (275.0,0.10155)
    (550.0,0.10142) (1375.0,0.10156) (2750.0,0.10150) (5500.0,0.10160)
};
\addlegendentry{4.8M samples seen}

\addplot[color=color5, mark=pentagon*, thick, mark size=1.8pt] coordinates {
    (5.5,0.11682) (55.0,0.11150) (110.0,0.11274) (275.0,0.11175)
    (550.0,0.11100) (1375.0,0.11202) (2750.0,0.11181) (5500.0,0.11117)
};
\addlegendentry{1.6M samples seen}

\end{axis}
\end{tikzpicture}
\end{minipage}%
\hfill
\begin{minipage}[t]{0.35\linewidth}
\vspace{0pt}
\captionof{figure}{\textbf{Data restriction ablation.} Validation loss at fixed training exposure (color) as the pool of unique footage available to the Base model is restricted from the full 5,500-hour corpus down to 5.5 hours. Loss degrades sharply only once repetition exceeds about 2,000 epochs (5.5h restriction); below that, additional unique footage brings diminishing returns for the Base model at the exposures tested here.}
\label{fig:data_restriction_ablation}
\end{minipage}
\vspace{-27pt}
\end{figure}
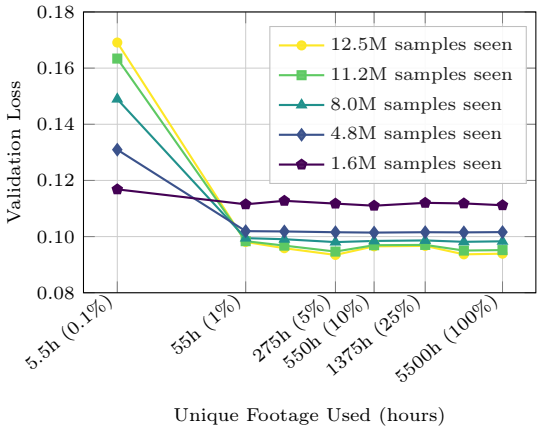

\autoref{fig:data_restriction_ablation} reports validation loss at five fixed exposure levels (1.6M to 12.5M samples seen) across this restriction spectrum. Loss stays nearly flat from 55 hours up to the full corpus: at $1.25\times10^7$ samples, a $100\times$ reduction in unique footage (5,500h to 55h) moves the loss only from 0.0939 to 0.0980. Only the extreme 5.5-hour restriction degrades sharply, where repetition reaches close to 2,000 epochs. Repetition can substitute for unique data, but only up to a point: below roughly 200 epochs, repeating a smaller pool of clips is harmless, consistent with observations that repeating data for a limited number of epochs matches the validation performance of fresh data in discrete diffusion models~\citep{prabhudesai2026diffusion,muennighoff2023scaling}. For the Base model at the exposures tested here, under 5 epochs over the full corpus, this limit is far off, so restricting unique footage costs little. Larger models trained for longer will need more epochs over the same corpus, moving closer to that limit, so genuinely new data should start to matter again.

\subsection{\texorpdfstring{\textbf{Compute scaling} ($x = C$)}{Compute scaling (x = C)}}\label{subsec:C}

\paragraph{\textbf{Goal.}} We investigate scaling under a fixed compute training budget \(C\) (TFLOPs), where model size \(N\) and training exposure \(D\) are jointly constrained. 
We assess the best achievable validation loss for a given compute budget and whether an optimal allocation between model capacity and training exposure emerges. 

\begin{figure}[t]
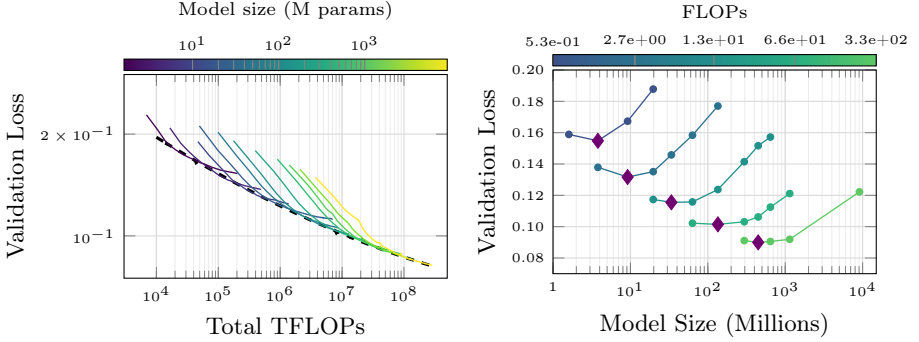

\centering
\begin{subfigure}{0.48\textwidth}
    \centering

    \caption{\textbf{Training curve envelope.} Per-model training curves (colored by parameter count) and the fitted compute-scaling law $L(C) = 0.0522 +  0.5955 \cdot C^{-0.154}$ (black dashed line), tracing the lower envelope of achievable validation loss versus compute.}
    \label{fig:scaling_compute_size_a}
\end{subfigure}
\hfill
\begin{subfigure}{0.48\textwidth}
    \centering
%
    \caption{\textbf{IsoFLOP.} Each dot is a fully converged model at a given model size, colored by its compute budget; purple diamonds mark the best-performing configuration among the tested model sizes and training exposures for each compute budget.}
    \label{fig:scaling_compute_size_b}
\end{subfigure}
\vspace{-1em}
\caption{\textbf{Compute scaling analysis.} Lower envelope of validation loss versus compute (a), and iso-FLOP optimal $(N,D)$ configurations for each budget (b).}
\label{fig:scaling_compute_size}
\vspace{-0.5cm}
\end{figure}

\paragraph{\textbf{Setup.}} We use approximately 200 trained models spanning a wide range of parameter counts and training exposures. We consider a series of fixed compute budgets and, for each budget, retain the best-performing configuration among all tested \((N,D)\) pairs. These Pareto-optimal points are then used to fit the compute scaling law using the same functional form as in \autoref{eq:scaling_law}:
\begin{equation}
\mathcal{L}(C) = 0.0522 + 0.5955 \, C^{-0.15443}. 
\label{eq:model_scaling_fit_C}
\end{equation}
\paragraph{\textbf{Results.}} 
The fitted law accurately captures the lower envelope of achievable validation losses across compute budgets. 
Increasing compute consistently improves performance (\autoref{fig:scaling_compute_size_a}), while the optimal allocation between model size and training exposure shifts toward larger models as the compute budget increases (\autoref{fig:scaling_compute_size_b}). The fitted asymptotic loss (\(L_0 \approx 0.0522\)) suggests further improvements remain possible with additional compute, although this estimate should be interpreted cautiously given the finite dataset size.

\begin{mdframed}[linewidth=0.8pt]
\textbf{Prediction for the 9B run.}
The compute scaling law in~\autoref{eq:model_scaling_fit_C} provides a unified description of performance across both model size and training exposure. Combining the predicted optimal compute allocation with the planned 9B architecture gives, for a model trained on approximately \(10^7\) samples (\(1.438 \times 10^9\) TFLOPs), an estimated validation loss of
\[
\mathcal{L}(1.438 \times 10^9) \approx 0.0753.
\]
This prediction is consistent with the model-size and training-exposure analyses, placing the planned 9B run close to the compute-optimal regime.
\end{mdframed}

\section{Scaling to 9B}
\label{sec:experiments}

Guided by the scaling laws, we train our 9B model and evaluate it against prominent open-source driving world models~\citep{hassan2025gem, gao2024vista}. We first verify its validation loss against our empirical predictions (\autoref{subsec:9b_val}) before assessing visual generation quality (\autoref{subsec:visual_quality}). To ensure a fair comparison with trajectory-controllable baselines, the model is augmented with trajectory conditioning (\autoref{para:traj}).

\subsection{Validation Loss at 9B scale}\label{subsec:9b_val}

\textbf{Scaling-law prediction.}
Across all three scaling regimes, the fitted laws consistently suggest that further scaling remains beneficial: model scaling shows no saturation, training dynamics remain stable across sizes, and compute scaling favors allocating additional compute budget to larger models. Based on these observations, we train the 9B model using the predicted compute-optimal configuration, corresponding to approximately \(10^7\) training samples. Before training, the scaling laws predicted a validation loss of \(0.0753\).

\textbf{Extrapolation accuracy.}
After training up to \(1.2 \times10^7\) samples, i.e., approximately 2 epochs, the 9B model reaches a validation loss of \(0.0781\), the lowest among all evaluated models, demonstrating continued improvements when scaling from 1B to 9B parameters. This closely matches the extrapolated prediction, with only a \(3.6\%\) relative error, despite extrapolating \(8\times\) beyond the largest model used for fitting (1.1B parameters) and without intermediate scales. These results show that scaling laws fitted below 1.1B remain accurate to a few percent in the 9B regime.

\textbf{Optimization at scale.}
The small residual gap between the predicted and observed losses is likely due to the need to adapt optimization strategies at larger scales. The scaling-law extrapolation assumes the same training recipe used for fitting, whereas stable optimization at 9B parameters required modifying the learning rate schedule. Following observations from \citep{esser2024scaling}, we reduced the learning rate from \(10^{-4}\) to \(10^{-5}\) after 3.2M samples to mitigate training instability. This adjustment is consistent with prior studies of large-scale video diffusion transformers~\citep{yin2025scaling}, which find such models highly sensitive to learning rate and batch size at scale. Thus, the deviation from the fitted recipe likely explains the small difference between the extrapolated and achieved losses. 

\begin{figure}[t]
\centering
\begin{minipage}[t]{0.62\linewidth}
\centering
\begin{tikzpicture}[baseline=(current bounding box.north)]
  \begin{axis}[
    width=\linewidth, height=0.65\linewidth,
    xmin=0.0, xmax=1.2,
    ymin=0.075, ymax=0.112,
    xlabel={Training exposure ($\times 10^{7}$ samples)},
    ylabel={Validation Loss},
    xtick={0.0, 0.2, 0.4, 0.6, 0.8, 1.0},
    xticklabels={0.0, 0.2, 0.4, 0.6, 0.8, 1.0},
    ytick={0.080, 0.085, 0.090, 0.095, 0.100, 0.105, 0.110},
    yticklabels={0.080, 0.085, 0.090, 0.095, 0.100, 0.105, 0.110},
    grid=both,
    grid style={line width=.2pt, draw=gray!15},
    major grid style={line width=.4pt, draw=gray!30},
    legend style={at={(0.97,0.97)}, anchor=north east, draw=gray!40, fill=white, font=\scriptsize, cells={anchor=west}},
    x tick label style={/pgf/number format/fixed, /pgf/number format/precision=1, font=\scriptsize},
    y tick label style={/pgf/number format/fixed, /pgf/number format/precision=3, font=\scriptsize},
    label style={font=\scriptsize},
    xlabel near ticks,
    ylabel near ticks,
    legend cell align={left},
    clip=true,
    clip mode=individual
  ]

    \addplot[color=pltblue, mark=*, mark size=1.0pt, line width=0.8pt] coordinates {
      (0.048,0.110233) (0.064,0.103087) (0.080,0.098672) (0.096,0.095951) (0.112,0.093986)
      (0.128,0.092148) (0.144,0.089909) (0.160,0.088293) (0.176,0.087752) (0.192,0.087105)
      (0.208,0.086857) (0.224,0.086359) (0.240,0.086057) (0.256,0.085682) (0.272,0.085470)
      (0.288,0.085043) (0.304,0.084618) (0.320,0.083878) (0.336,0.083160) (0.352,0.082646)
      (0.368,0.082150) (0.384,0.082131) (0.400,0.081962) (0.416,0.081897) (0.432,0.081625)
      (0.448,0.081423) (0.448,0.081349) (0.480,0.081181) (0.496,0.081171) (0.512,0.081110)
      (0.528,0.081237) (0.544,0.080898) (0.560,0.080682) (0.576,0.080565) (0.592,0.080355)
      (0.608,0.080337) (0.624,0.080146) (0.640,0.080149) (0.656,0.083479) (0.672,0.080066)
      (0.688,0.079946) (0.704,0.080119) (0.720,0.079712) (0.736,0.079604) (0.752,0.079607)
      (0.768,0.079663) (0.784,0.079476) (0.800,0.079297) (0.816,0.079135) (0.832,0.079137)
      (0.848,0.079156) (0.864,0.079007) (0.880,0.078955) (0.896,0.078887) (0.912,0.078834)
      (0.928,0.078631) (0.944,0.078742) (0.960,0.078616) (0.976,0.078517) (0.992,0.078596)
      (1.008,0.078542) (1.024,0.078793) (1.040,0.078476) (1.056,0.078390) (1.072,0.078309)
      (1.088,0.078198) (1.104,0.078246) (1.120,0.078545) (1.136,0.078145)
    };
    \addlegendentry{Measured}

    \addplot[color=black, line width=1.0pt, no marks] coordinates {
      (0.048,0.109922) (0.064,0.103261) (0.080,0.098978) (0.096,0.095961) (0.112,0.093706)
      (0.128,0.091948) (0.144,0.090533) (0.160,0.089367) (0.176,0.088387) (0.192,0.087549)
      (0.208,0.086825) (0.224,0.086191) (0.240,0.085630) (0.256,0.085131) (0.272,0.084683)
      (0.288,0.084279) (0.304,0.083912) (0.320,0.083576) (0.336,0.083269) (0.352,0.082986)
      (0.368,0.082724) (0.384,0.082481) (0.400,0.082255) (0.416,0.082045) (0.432,0.081847)
      (0.448,0.081663) (0.464,0.081489) (0.480,0.081325) (0.496,0.081170) (0.512,0.081024)
      (0.528,0.080886) (0.544,0.080754) (0.560,0.080630) (0.576,0.080511) (0.592,0.080397)
      (0.608,0.080289) (0.624,0.080186) (0.640,0.080087) (0.656,0.079993) (0.672,0.079902)
      (0.688,0.079815) (0.704,0.079731) (0.720,0.079651) (0.736,0.079574) (0.752,0.079499)
      (0.768,0.079428) (0.784,0.079358) (0.800,0.079291) (0.816,0.079227) (0.832,0.079165)
      (0.848,0.079104) (0.864,0.079046) (0.880,0.078989) (0.896,0.078934) (0.912,0.078812)
      (0.928,0.078830) (0.944,0.078780) (0.960,0.078731) (0.976,0.078684) (0.992,0.078638)
      (1.008,0.078593) (1.024,0.078550) (1.040,0.078508) (1.056,0.078466) (1.072,0.078426)
      (1.088,0.078387) (1.104,0.078349) (1.120,0.078312) (1.136,0.078276)
    };
    \addlegendentry{Estimated scaling law}

    \addplot[color=purple!80!black, fill=purple, mark=diamond*, mark size=3.0pt, only marks] coordinates {
      (1.136,0.078145)
    };
    \addlegendentry{Best measured loss: 0.078145}

  \end{axis}
\end{tikzpicture}
\end{minipage}%
\hfill
\begin{minipage}[t]{0.35\linewidth}
\vspace{0pt}
\captionof{figure}{\textbf{Training-scaling fit for the 9B model.} Validation loss during training and the corresponding power law fit ($\mathcal{L}(D) = 0.074799 + 0.00381\, D^{-0.7309}$), which suggests that additional training could still be beneficial to narrow the gap to the asymptotic loss $L_0$.}
\label{fig:9b_fit}
\end{minipage}
\vspace{-10pt}
\end{figure}
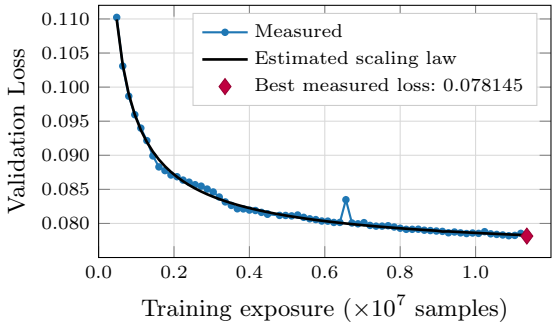

\textbf{Scaling efficiency at 9B.}
Nevertheless, the achieved loss further highlights the benefits of increasing model capacity. The 9B model reaches a loss comparable to the fitted asymptote of the 1.1B model (\(L_0=0.078\), \autoref{tab:scaling_laws_training_exp}), a level that the smaller 1.1B model would require more than an order of magnitude additional training exposure to approach. In contrast, the 9B model reaches this regime within \(1.2 \times10^7\) samples while remaining far from its own asymptote (\autoref{fig:9b_fit}). Moreover, the post-hoc fitted asymptotic loss (\(L_0 \approx 0.0748\)) sits within 1\% of the a-priori compute-law prediction of 0.0753, providing an independent confirmation of the extrapolated estimate.

\subsection{Generation Quality}\label{subsec:visual_quality}

\paragraph{\textbf{Goal.}} Beyond validation loss, we assess whether scaling also improves perceptual generation quality and trajectory controllability, and how the resulting models compare to existing driving world models.

\paragraph{\textbf{Setup.}}
\textit{Trajectory conditioning.}\label{para:traj} Lacking ground-truth ego-trajectories, we extract pseudo-trajectories with OccAny~\citep{cao2026occany}, which attains $0.90$m ADE on the Vista nuScenes val.\ set~\cite{gao2024vista}, ahead of GEM's pipeline~\cite{hassan2025gem} ($1.63$m) and DA3~\cite{lin2026depthanything3} ($1.12$m). Since the pseudo-labels are imperfect and the dataset is biased toward driving straight, we refine them as follows: (i) a kinematic filter that drops physically implausible segments; and (ii) a scenario-balanced sampler over eleven ego-action classes (idle, weak/strong forward, gentle/hard braking, weak/strong left/right turns, backward straight/turning) that up-weights rare maneuvers.

Each trajectory is $25$ bird's-eye-view waypoints in meters relative to the first frame, injected via AdaLN (\autoref{fig:arch}). To match Wan's temporal compression (first frame kept, subsequent groups of four compressed into one), we group waypoints identically: after per-coordinate sinusoidal embedding, the first waypoint uses a zero embedding and the rest are concatenated in groups of four, each mapped by an MLP to the scale, shift modulations of every block and added to the per-frame timestep embedding. For CFG~\cite{ho2022classifier}, we drop the trajectory with probability $0.15$. 

\begin{table}[t]
\centering
\caption{\textbf{Scaling Model size on 2.5\,s video generation quality on NATIX}, conditioned on a single frame. Lower values indicate better performance; best results in each column are bold. Fr\'echet distances are computed 
with different backbones. Models marked with * use trajectory-conditioning post-training. Green subscripts indicate the ADE 
boost over 
the 
corresponding unconditioned model. Larger models consistently improve generation quality, consistent with lower pre-training validation loss. }
\label{tab:fid_fvd_scaling}
\vspace{-10pt}
\begin{tabular}{lccccc}
\toprule
Model & FID$_{\text{Inception}}$ $\downarrow$ & FID$_{\text{DINO}}$ $\downarrow$ & FVD$_{\text{I3D}}$ $\downarrow$ & FVD$_{\text{VideoMAE}}$ $\downarrow$ & ADE $\downarrow$ \\
\midrule
Tiny  & 26.87 & 280.19 & 175.99 & 198.40 & -- \\
Base  & 10.66 & 132.51 & 68.43  & 116.94 & -- \\
Large & 8.67 & 117.76 & 48.09  & 99.43  & -- \\
1B    & 5.61  & 87.05  & \textbf{33.94} & 89.49 & -- \\
9B    & \textbf{4.91} & \textbf{60.81} & 37.16 & \textbf{75.86} & -- \\ \midrule
1B*   & 4.41  & 63.93  & 32.86  &  49.28 & 3.92$_{\textcolor{ForestGreen}{-0.98}}$ \\
9B*   &  \textbf{4.03}  & \textbf{42.16}  &  \textbf{24.68}  &  \textbf{44.95} & 3.84$_{\textcolor{ForestGreen}{-0.18}}$ \\
\bottomrule
\end{tabular}
\vspace{-12pt}
\end{table}

\textit{Post-training.} We fine-tune the full backbone, initialized from the corresponding (non-trajectory) checkpoints, using the aforementioned scenario-balanced sampler over eleven ego-action classes.
While the 1B model is optimized directly for $100$K iterations, the 9B model requires a two-stage schedule because the timestep embedding dominates trajectory modulation. 
We first train only the trajectory MLP for $60$K iterations at a learning rate of $10^{-3}$, followed by $120$K iterations of joint fine-tuning using $10^{-5}$ for the backbone and $10^{-4}$ for the trajectory MLP.

\textit{Metrics.} We evaluate visual fidelity, temporal consistency, and trajectory accuracy. We report FID~\citep{heusel2017gans} with two image encoders (Inception~\citep{szegedy2016rethinking} and DINOv2~\citep{oquab2024dinov2}) for per-frame fidelity, and FVD~\citep{unterthiner2019fvd} with two video encoders (I3D~\citep{carreira2017quo} and VideoMAE~\citep{tong2022videomae}) for temporal consistency, the paired encoders giving complementary views in each case. Finally, we report the Average Displacement Error (ADE) between the ego-trajectory of the real clip and the one re-extracted from the generated video, both obtained with OccAny~\citep{cao2026occany}.

\textit{Test set.} The evaluation is performed on a filtered test set of 2,000 videos, uniformly balanced across eleven ego-action classes (idle, weak/strong forward, gentle/hard braking, weak/strong left/right turns, backward straight/turning), ensuring consistent coverage of ego-motion modes. All models are evaluated in a 1-frame conditioning setting, generating 2.5-second videos at 9 Hz.

\begin{table}[t]
\centering
\caption{\textbf{Comparison on the nuScenes benchmark.} Quantitative evaluation of 2.5\,s video generation conditioned on a single input frame (lower is better; ``--'' indicates the metric is not reported). Results are evaluated on two test splits matching the protocols of Vista (5,369 sequences) and Epona (1,690 sequences). Our models achieve state-of-the-art performance, demonstrating the benefit of scaling both model size and training data, followed by domain-specific fine-tuning on nuScenes.}
\vspace{-10pt}
\begin{tabular}{lcccc}
\toprule
Model & FID$_{\text{Inception}}$ $\downarrow$ & FID$_{\text{DINO}}$ $\downarrow$ & FVD$_{\text{I3D}}$ $\downarrow$ & FVD$_{\text{VideoMAE}}$ $\downarrow$ \\
\midrule
DriveDreamer-2~\citep{zhao2025drivedreamer2}        & 25.0 & -- & 105.1 & -- \\
Drive-WM~\citep{wang2024driving}              & 15.8 & -- & 122.7 & -- \\
GenAD~\citep{yang2024genad}                 & 15.4 & -- & 184.0 & -- \\
Vista~\citep{gao2024vista}                 & 6.9  & -- & 89.4  & -- \\
GEM~\citep{hassan2025gem}                   & 10.5 & -- & 158.5 & -- \\
Driving World~\citep{hu2024drivingworld}         & 7.4  & -- & 90.9  & -- \\
Epona~\citep{zhang2025epona}                 & 7.5  & -- & 82.8  & -- \\ 
\midrule
1B (ours) Vista split & 3.10 & 132.86 & 28.46 & \textbf{30.69} \\
9B (ours) Vista split & \textbf{2.72} & \textbf{87.12} &  \textbf{25.50} &  32.58 \\ \midrule
1B (ours) Epona split  & 3.83 & 137.20 & 34.55 & \textbf{38.00} \\
9B (ours) Epona split  & \textbf{3.61} & \textbf{92.73} & \textbf{31.52}  & 39.75 \\
\bottomrule
\end{tabular}
\label{tab:fid_fvd_nuscenes}
\vspace{-15pt}
\end{table}

\begin{figure}[t]
\centering
\begin{tikzpicture}
  \node[inner sep=0] (img) {\includegraphics[width=0.9\linewidth]{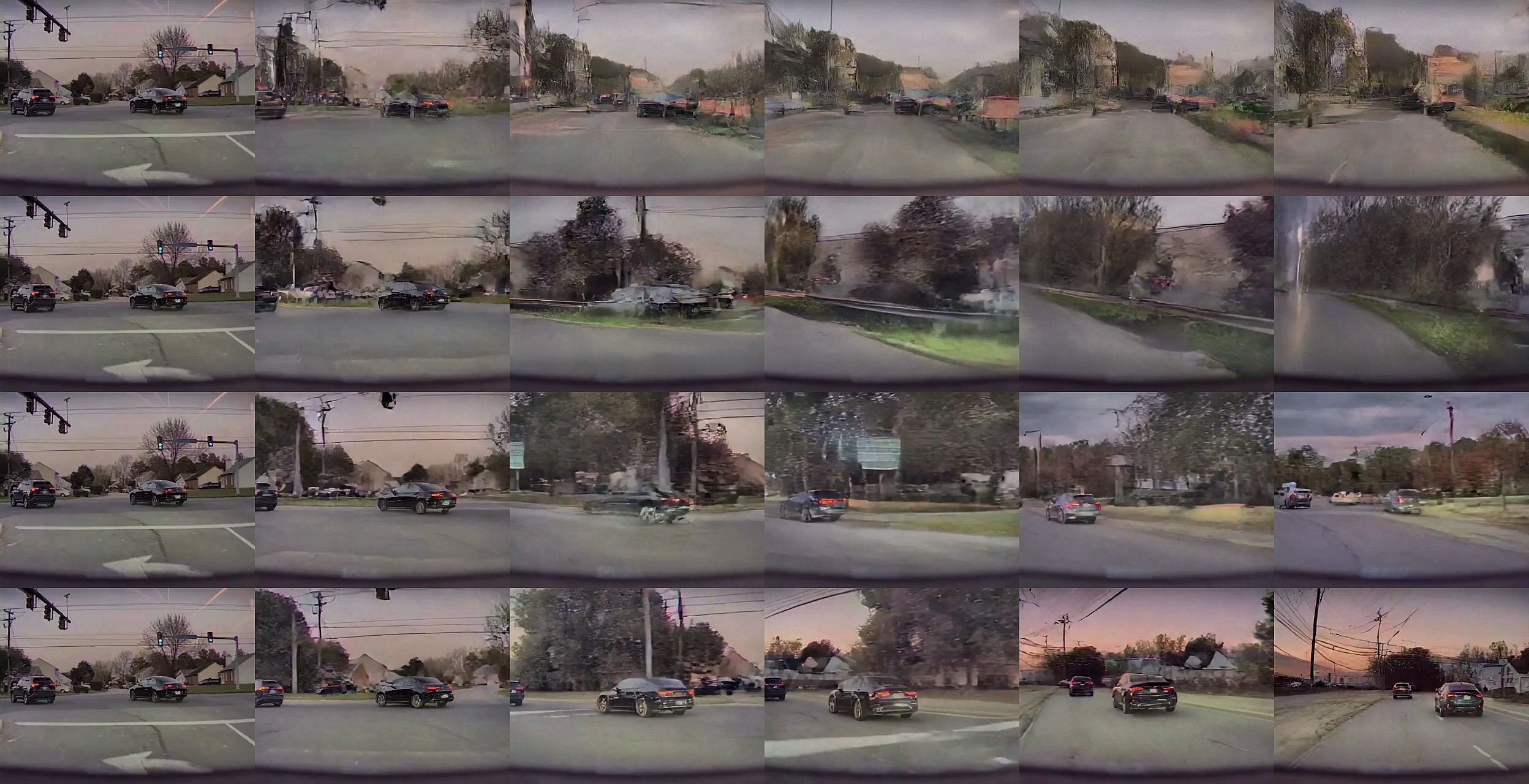}};

  \node[left=4pt] at ($(img.north west)!0.125!(img.south west)$) {\small 19M};
  \node[left=4pt] at ($(img.north west)!0.375!(img.south west)$) {\small 135M};
  \node[left=4pt] at ($(img.north west)!0.625!(img.south west)$) {\small 1.1B};
  \node[left=4pt] at ($(img.north west)!0.875!(img.south west)$) {\small 9.0B};

  \draw[->, thick]
    ($(img.south west)+(0,-2mm)$) --
    ($(img.south east)+(0,-2mm)$)
    node[midway, below=4pt] {\small Time (0s $\rightarrow$ 5s)};
\end{tikzpicture}
\vspace{-2.5em}
\caption{\textbf{Visual Quality on 5-second generations.} Small models (<20M parameters) capture the overall scene layout but quickly lose object consistency across frames. The base model (135M) better understands the scene, e.g., it correctly predicts that the car is turning left, but struggles to maintain coherent dynamics over longer horizons. Larger models (1.1B parameters) further improve visual quality and scene structure, yet still exhibit temporal inconsistencies and break down after approximately 2.5 seconds. Only our largest 9B models accurately capture the scene dynamics while maintaining high visual fidelity and temporal coherence throughout the entire rollout.}
\label{fig:vizu_scaling_model}
\vspace{-0.5cm}
\end{figure}
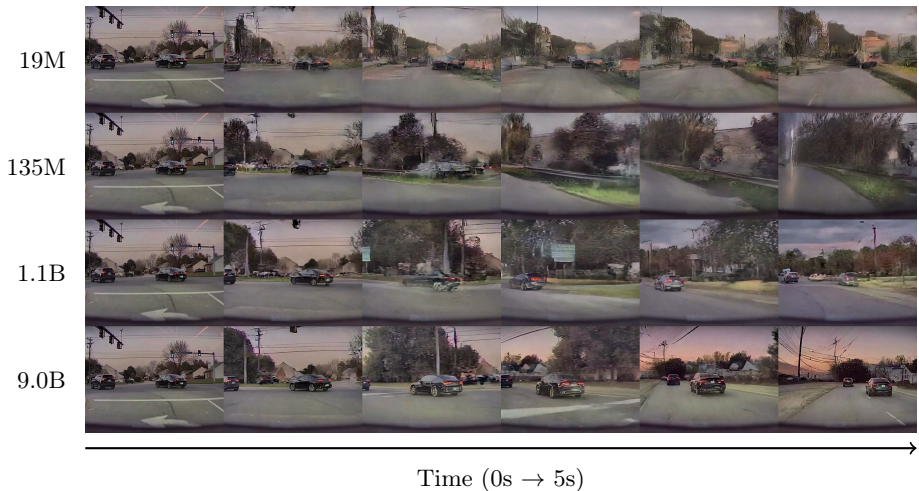

\paragraph{\textbf{Results.}} Scaling model size improves evaluation metrics, as shown in~\autoref{tab:fid_fvd_scaling}, with the largest gains in video-based perceptual scores, indicating better temporal coherence as capacity increases. The jump from 1B to 9B yields diminishing but consistent improvements in visual fidelity, with the strongest gains observed on video metrics, suggesting improved long-range motion consistency, as explicitly shown on a hard case in~\autoref{fig:vizu_scaling_model}. Trajectory fine-tuning (denoted *) improves the perceptual metrics further and reduces the ADE (\autoref{tab:fid_fvd_scaling}).

\textit{Comparison with prior work.} For a direct comparison with existing methods (\autoref{tab:fid_fvd_nuscenes}), we finetune the 1B model (learning rate $10^{-4}$) and the 9B model (learning rate $10^{-4}$ for the trajectory MLP, $10^{-5}$ for the rest) on nuScenes \citep{caesar2020nuscenes} for $25$K iterations. Our approach achieves the best performance on both image- and video-based perceptual metrics. We attribute this to two factors: large-scale pre-training on diverse driving scenes provides a strong prior that transfers effectively, and, unlike GEM and Vista, which use parameter-efficient adaptation, we fully fine-tune all parameters. Note that the 9B–1B gap is smaller on nuScenes than on NATIX, and the 9B even underperforms on FVD$_\text{VideoMAE}$, likely as its $10\times$ smaller learning rate weakens trajectory adherence at equal iterations.

\section{Conclusion}
\vspace{-5pt}
We hope these scaling laws provide a practical basis for training future driving world models.
This work answers its title question: 5,500 hours of driving take a from-scratch video diffusion model to a new open-source state-of-the-art for driving video generation, as measured on nuScenes.
Across more than 200 training runs spanning 1.6M to 1.1B parameters, three findings emerge.
(1) The training-exposure exponent is far steeper than the model-size exponent, so longer training is the fastest way to improve a fixed model, while compute-optimal budgets still favor larger models as capacity keeps lowering the asymptotic loss.
(2) The fixed corpus does not break the laws: repeated data behaves like fresh samples, so the 5,500 hours are not yet the bottleneck, which we hypothesize follows from flow matching resampling noise and timestep at every pass, unlike the fixed targets of autoregressive training.
(3) The laws extrapolate: a 9B-parameter model, to our knowledge the largest open-source video diffusion model trained from scratch on driving data, reaches a validation loss predicted before training to within 3.6\% of a law extrapolated 8$\times$ beyond the largest fitted model, the lowest loss among all our models.
The main limitation is the scale gap between the models used for fitting and the flagship: intermediate scales (e.g., 3B--4B) would tighten extrapolated estimates.
\\
\textbf{Limitations.} While our models are trained from scratch, extending the scaling analysis to jointly account for the frozen VAE is a promising direction. Similarly, our scaling laws assume a uniform learning rate; since the 9B model required adjusting the learning rate during training for stability, a hyperparameter grid search could further tighten our predictions.
\\
{\footnotesize
\textbf{Acknowledgment.}
This work was granted access to the HPC resources of IDRIS under the allocation A0201016203 made by GENCI. We acknowledge the EuroHPC Joint Undertaking for awarding the project IDs EHPC-AIF-2026FL01-008, EHPC-AIF-2026FL01-258, and EHPC-AIF-2026LS11-001 access to the EuroHPC supercomputer MareNostrum~5, hosted by the Barcelona Supercomputing Center (BSC), Spain. This work was supported by the European Union's Horizon Europe research and innovation programme under grant agreement No.~101214398 (ELLIOT). The authors would like to thank Florent Bartoccioni for his assistance with data management, Valentin Gerard for his contributions to the diffusion model and auto-encoder selection, and Mustafa Shukor for his help in establishing the scaling laws.
\par}

\bibliographystyle{splncs04}
\bibliography{ref}

\end{document}